\documentclass[10pt,twocolumn,letterpaper]{article}
\usepackage{iccv}
\usepackage{times}
\usepackage{epsfig}
\usepackage{graphicx}
\usepackage{amsmath}
\usepackage{amssymb}
\usepackage{dsfont}
\usepackage[pagebackref=true,breaklinks=true,letterpaper=true,colorlinks,bookmarks=false]{hyperref}

\usepackage[textwidth=1.8cm]{todonotes}
\usepackage{enumitem}
\usepackage{amsmath}
\usepackage{adjustbox}
\usepackage{listings}
\usepackage{times}
\usepackage{epsfig}
\usepackage{graphicx}
\usepackage{amsmath}
\usepackage{amssymb}
\usepackage{wrapfig}
\usepackage{caption}
\usepackage{subcaption}
\usepackage{float}
\usepackage{array}
\usepackage{color}
\usepackage{multirow}
\usepackage{diagbox}
\usepackage{booktabs}
\usepackage{siunitx}
\usepackage{color}
\usepackage{graphbox}
\usepackage{cuted}
\usepackage[table]{}
\usepackage[accsupp]{axessibility}  % Improves PDF readability for those with disabilities.

\usepackage[normalem]{ulem}
\usepackage{tikz}
\usepackage{multirow}
\usepackage{varwidth}
\usepackage{pifont} %
\usepackage{enumitem}
\definecolor{pink}{RGB}{213,126,190}
\definecolor{orange}{RGB}{239,133,54}
\definecolor{lightblue}{RGB}{173,216,230}
\definecolor{darkblue}{RGB}{8,81,156}
\definecolor{stanfordgrey}{RGB}{46,45,41}
\definecolor{cardinalred}{RGB}{253,141,60}
\definecolor{codegreen}{rgb}{0,0.4,0}
\definecolor{codegray}{rgb}{1.0,0.5,0.5}
\definecolor{codepurple}{rgb}{0.58,0,0}
\definecolor{tealblue}{rgb}{0,0.5,0.5}
\definecolor{codebackcolour}{rgb}{0.95,0.95,0.92}
\definecolor{darkgreen}{RGB}{0,127,0}
\definecolor{yellow}{RGB}{240,164,47}
\definecolor{darkred}{RGB}{200,0,0}
\definecolor{orange}{rgb}{1,0.5,0}

\def\greencheckmark{\textcolor{darkgreen}{\checkmark}}
\def\redxmark{\textcolor{darkred}{\ding{55}}}  %

\usepackage[capitalize]{cleveref}
\crefname{section}{Sec.}{Secs.}
\Crefname{section}{Section}{Sections}
\Crefname{table}{Table}{Tables}
\crefname{table}{Tab.}{Tabs.}

\iccvfinalcopy % *** Uncomment this line for the final submission

\def\iccvPaperID{12782} % *** Enter the ICCV Paper ID here

\ificcvfinal\fi
\usepackage{xr}
\begin{document}
%%%%%%%%% TITLE
\title{NeO 360: Neural Fields for Sparse View Synthesis 
of Outdoor Scenes}

\author{Muhammad Zubair Irshad$^1$$^2$\hspace{.2cm} Sergey Zakharov$^2$\hspace{.2cm} Katherine Liu$^2$\hspace{.2cm} Vitor Guizilini$^2$\hspace{.2cm} Thomas Kollar$^2$\\ 
Adrien Gaidon$^2$\hspace{.2cm} Zsolt Kira*$^1$\hspace{.2cm} Rares Ambrus*$^2$\\[0.01cm]
\footnotesize{* denotes shared last authorship}\\[0.09cm]
$^1$Georgia Institute of Technology\hspace{.4cm} $^2$Toyota Research Institute\\[0.01cm]
{\tt\footnotesize \{mirshad7, zkira\}@gatech.edu, \{firstname.lastname\}@tri.global}
}

\newcommand{\titleShort}{NeO 360}
% \maketitle

\twocolumn[{%
\renewcommand\twocolumn[1][]{#1}%
\maketitle

\begin{center}
    \centering
    \includegraphics[width=0.98\textwidth]{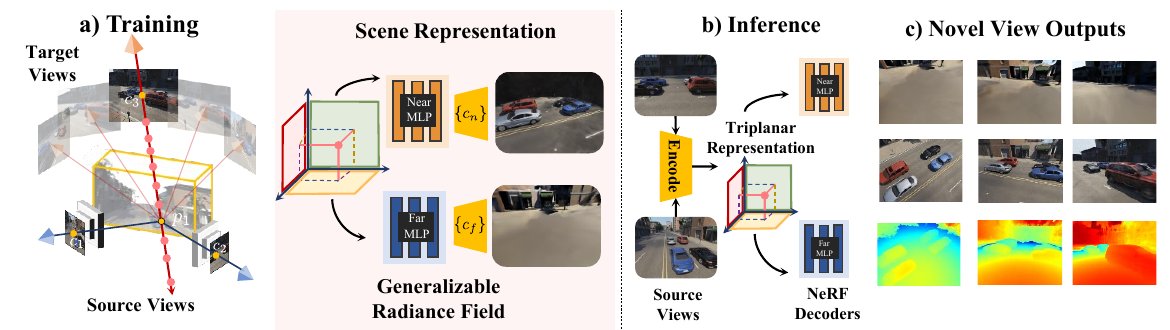}
    \captionsetup{width=\linewidth}
    \captionof{figure}{
    \textbf{Overview:} a) Given just a single or a few input images from a novel scene, our method renders new 360$^{\circ}$ views of complex unbounded outdoor scenes b) We achieve this by constructing an image-conditional triplane representation to model the 3D surrounding from various perspectives. c) Our method \textit{generalizes} across novel scenes and viewpoints for complex 360$^{\circ}$ outdoor scenes.
    }
   \label{fig:teaser}
\end{center}%
}]
\ificcvfinal\thispagestyle{empty}\fi

%%%%%%%%% BODY TEXT
\begin{abstract}
    \vspace{-5pt}
    Recent implicit neural representations have shown great results for novel view synthesis. However, existing methods require expensive per-scene optimization from many views hence limiting their application to real-world unbounded urban settings where the objects of interest or backgrounds are observed from very few views. To mitigate this challenge, we introduce a new approach called \titleShort{}, Neural fields for sparse view synthesis of outdoor scenes. \titleShort{} is a generalizable method that reconstructs 360$^{\circ}$ scenes from a single or a few posed RGB images. The essence of our approach is in capturing the distribution of complex real-world outdoor 3D scenes and using a hybrid image-conditional triplanar representation that can be queried from any world point. Our representation combines the best of both voxel-based and bird’s-eye-view (BEV) representations and is more effective and expressive than each. \titleShort{}'s representation allows us to learn from a large collection of unbounded 3D scenes while offering generalizability to new views and novel scenes from as few as a single image during inference. We demonstrate our approach on the proposed challenging 360$^{\circ}$ unbounded dataset, called NeRDS 360, and show that~\titleShort{} outperforms state-of-the-art generalizable methods for novel view synthesis while also offering editing and composition capabilities. Project page:
\href{https://zubair-irshad.github.io/projects/neo360.html}{zubair-irshad.github.io/projects/neo360.html}
\end{abstract}
\vspace{-10pt}
\section{Introduction}
\label{sec:intro}
\looseness=-1
Advances in learning-based implicit neural representations have demonstrated promising results for high-fidelity novel-view synthesis~\cite{xie2021neural}, doing so from multi-view images~\cite{ueda2022neural,wang2021neus}. The capability to infer accurate 3D scene representations has benefits in autonomous driving~\cite{ost2021neural, fu2022panoptic} and robotics~\cite{irshad2022shapo, Ortiz:etal:iSDF2022, irshad2022centersnap}.

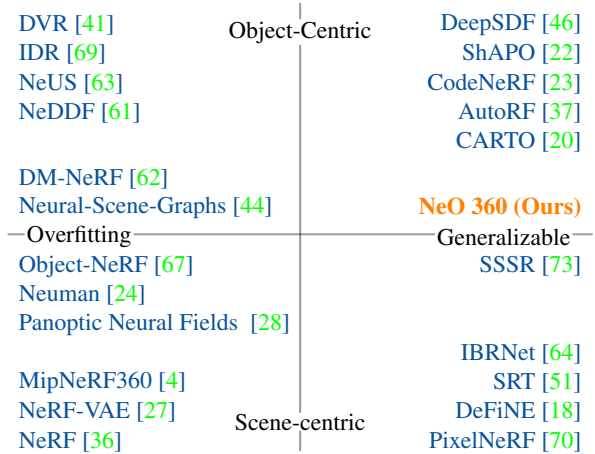
\begin{figure}[!t]
  \centering
  \resizebox{0.95\columnwidth}{!}{\begin{tikzpicture}
\draw[ line width=0.2 mm, color=black, opacity=0.4] (3,0) -- (3,4.6);
\draw[ line width=0.2 mm, color=black, opacity=0.4] (0,2.225) -- (6,2.225);

\node [ color=darkblue, rotate=0, anchor=west, fill=white,rounded corners=2pt,inner sep=0.2pt] at (0.1,2.8) {\scriptsize DM-NeRF~\cite{wang2022dm}};
\node [ color=darkblue, rotate=0, anchor=west, fill=white,rounded corners=2pt,inner sep=0.2pt] at (0.1,2.5) {\scriptsize Neural-Scene-Graphs~\cite{ost2020neural}};
\node [ color=darkblue, rotate=0, anchor=west, fill=white,rounded corners=2pt,inner sep=0.2pt] at (0.1,1.9) 
{\scriptsize Object-NeRF~\cite{yang2021objectnerf}};
\node [ color=darkblue, rotate=0, anchor=west, fill=white,rounded corners=2pt,inner sep=0.2pt] at (0.1,1.6) {\scriptsize Neuman~\cite{neural-human-radiance-field}};
\node [ color=darkblue, rotate=0, anchor=west, fill=white,rounded corners=2pt,inner sep=0.2pt] at (0.1,1.3) {\scriptsize Panoptic Neural Fields ~\cite{kundu2022panoptic}};

\node [ color=darkblue, rotate=0, anchor=west, fill=white,rounded corners=2pt,inner sep=0.2pt] at (0.1,0.1) {\scriptsize NeRF~\cite{mildenhall2020nerf}};
\node [ color=darkblue, rotate=0, anchor=west, fill=white,rounded corners=2pt,inner sep=0.2pt] at (0.1,0.4) {\scriptsize NeRF-VAE~\cite{kosiorek2021nerf}};

\node [ color=darkblue, rotate=0, anchor=west, fill=white,rounded corners=2pt,inner sep=0.2pt] at (0.1,0.7) {\scriptsize MipNeRF360~\cite{barron2022mip}};

\node [ color=darkblue, rotate=0, anchor=west, fill=white,rounded corners=2pt,inner sep=0.2pt] at (0.1,4.375) {\scriptsize DVR~\cite{niemeyer2020differentiable}};
\node [ color=darkblue, rotate=0, anchor=west, fill=white,rounded corners=2pt,inner sep=0.2pt] at (0.1,4.075) {\scriptsize IDR~\cite{yariv2020multiview}};
\node [ color=darkblue, rotate=0, anchor=west, fill=white,rounded corners=2pt,inner sep=0.2pt] at (0.1,3.775) {\scriptsize NeUS~\cite{wang2021neus}};
\node [ color=darkblue, rotate=0, anchor=west, fill=white,rounded corners=2pt,inner sep=0.2pt] at (0.1,3.475) {\scriptsize NeDDF~\cite{ueda2022neural}};

\node [ color=black, rotate=0, anchor=east, fill=white,rounded corners=2pt,inner sep=0.6pt] at (5.8,2.2) {\scriptsize Generalizable};

\node [ color=black, rotate=0, anchor=east, fill=white,rounded corners=2pt,inner sep=0.6pt] at (1.3,2.2) {\scriptsize Overfitting};

\node [color=black, rotate=0, anchor=center, fill=white,rounded corners=2pt,inner sep=0.2pt] at (3.0,0.3) {\scriptsize Scene-centric};

\node [ color=orange, rotate=0, anchor=east, fill=white,rounded corners=2pt,inner sep=0.2pt] at (5.9,2.5) {\scriptsize \textbf{\titleShort{} (Ours)}};
\node [ color=darkblue, rotate=0, anchor=east, fill=white,rounded corners=2pt,inner sep=0.2pt] at (5.9,1.9) {\scriptsize SSSR~\cite{zakharov2021single}};

\node [ color=darkblue, rotate=0, anchor=east, fill=white,rounded corners=2pt,inner sep=0.2pt] at (5.9,0.1) {\scriptsize PixelNeRF~\cite{yu2020pixelnerf}};
\node [ color=darkblue, rotate=0, anchor=east, fill=white,rounded corners=2pt,inner sep=0.2pt] at (5.9,0.4) {\scriptsize DeFiNE~\cite{guizilini2022depth}};
\node [ color=darkblue, rotate=0, anchor=east, fill=white,rounded corners=2pt,inner sep=0.2pt] at (5.9,0.7) {\scriptsize SRT~\cite{srt22}};
\node [ color=darkblue, rotate=0, anchor=east, fill=white,rounded corners=2pt,inner sep=0.2pt] at (5.9,1.0) {\scriptsize IBRNet~\cite{wang2021ibrnet}};

\node [ color=black, rotate=0, anchor=center, fill=white,rounded corners=2pt,inner sep=0.6pt] at (3,4.3) {\scriptsize Object-Centric};

\node [ color=darkblue, rotate=0, anchor=east, fill=white,rounded corners=2pt,inner sep=0.2pt] at (5.9,4.375) {\scriptsize DeepSDF~\cite{park2019deepsdf}};

\node [ color=darkblue, rotate=0, anchor=east, fill=white,rounded corners=2pt,inner sep=0.2pt] at (5.9,4.075) {\scriptsize ShAPO~\cite{irshad2022shapo}};

\node [ color=darkblue, rotate=0, anchor=east, fill=white,rounded corners=2pt,inner sep=0.2pt] at (5.9,3.775) {\scriptsize CodeNeRF~\cite{jang2021codenerf}};
\node [ color=darkblue, rotate=0, anchor=east, fill=white,rounded corners=2pt,inner sep=0.2pt] at (5.9,3.475) {\scriptsize AutoRF~\cite{mueller2022autorf}};

\node [ color=darkblue, rotate=0, anchor=east, fill=white,rounded corners=2pt,inner sep=0.2pt] at (5.9,3.175) {\scriptsize CARTO~\cite{heppert2023carto}};

\end{tikzpicture}}
  \caption{\textbf{Taxonomy of implicit representation methods} reconstructing appearance and shapes. The~$x$ and~$y$ correspond to (Generalizable vs Overfitting) and (Object-Centric vs Scene-Centric) dimensions, as discussed in Section~\ref{sec:intro}.
  } 
  \label{fig:related_work}
\vspace{-4mm}
\end{figure}

\looseness=-1
Despite great progress in neural fields~\cite{tewari2021advances} for indoor novel view synthesis, these techniques are limited in their ability to represent complex urban scenes as well as decompose scenes for reconstruction and editing. Specifically, previous formulations~\cite{boss2020nerd, park2020nerfies, zhang2020nerf++} have focused on per-scene optimization from a large number of views, thus increasing their computational complexity. This requirement limits their application to complex scenarios such as data captured by a moving vehicle where the geometry of interest is observed in just a few views. Another line of work focuses on object reconstructions~\cite{mueller2022autorf, 
irshad2022shapo} from single-view RGB~(Fig.~\ref{fig:related_work}). However, these approaches require accurate panoptic segmentation and 3D bounding boxes as input which is a strong supervisory signal and consists of multi-stage pipelines that can lead to error-compounding.

\looseness=-1
To avoid the challenge of acquiring denser input views of a novel scene in order to obtain an accurate 3D representation as well as the computational expense of per-scene optimization from many views, we propose to infer the representation of 360$^{\circ}$~unbounded scenes from just a single or a few posed RGB images of a novel outdoor environment. 

As shown in Fig.~\ref{fig:teaser}, our approach extends the NeRF++~\cite{zhang2020nerf++} formulation by making it generalizable. At the core of our method are  local features represented in the form of triplanes~\cite{chan2022efficient}. This representation is constructed as three perpendicular cross-planes, where each plane
models the 3D surroundings from one perspective and by merging
them, a thorough description of the 3D scene can be achieved. ~\titleShort{}'s image-conditional tri-planar representation efficiently encodes information from image-level features while offering a compact queryable representation for any world point. We use these features combined with the residual local image-level features to optimize multiple unbounded 3D scenes from a large collection of images. \titleShort{}'s 3D scene representation can build a strong prior for complete 3D scenes, which enables efficient 360$^{\circ}$ novel view synthesis for outdoor scenes from just a few posed RGB images. 

\begin{figure}[t]
    \centering
    \resizebox{0.9\linewidth}{!}{%
\includegraphics{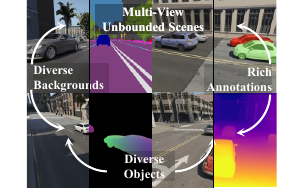}
    }
    \caption{Samples from our large scale \textbf{NeRDS 360}:~"\textbf{Ne}RF for~\textbf{R}econstruction,~\textbf{D}ecomposition and Scene~\textbf{S}ynthesis of 360$^{\circ}$ outdoor scenes" dataset comprising 75 unbounded scenes with full multi-view annotations and diverse scenes.
    }
    \label{fig:dataset_fig1}
    \vspace{-2mm}
\end{figure}%

To enable building a strong prior representation of unbounded outdoor scenes and given the scarcity of available multi-view data to train methods like NeRF, we also present a new large scale 360$^{\circ}$ unbounded dataset~(Figure~\ref{fig:dataset_fig1}) comprising of more than 70 scenes across 3 different maps. We demonstrate our approach's effectiveness on this challenging multi-view unbounded dataset in both few-shot novel-view synthesis and prior-based sampling tasks. In addition to learning a strong 3D representation for complete scenes, our method also allows for inference-time pruning of rays using 3D ground truth bounding boxes, thus enabling compositional scene synthesis from a few input views. In summary, we make the following contributions:
\begin{itemize}[noitemsep]
   \item{\textbf{A generalizable NeRF  architecture for outdoor scenes} based on tri-planar representation to extend the NeRF formulation for effective few-shot novel view synthesis of 360$^{\circ}$ unbounded environments.}
    \item{ \textbf{A large scale synthetic 360$^{\circ}$ dataset, called NeRDS 360, for 3D urban scene understanding} comprising multiple objects, capturing high-fidelity outdoor scenes with dense camera viewpoint annotations.}
    \item{Our proposed approach significantly outperforms all baselines for few-shot novel view synthesis on the NeRDS 360 dataset, showing 1.89 PNSR and 0.11 SSIM absolute improvement number for the 3-view novel-view synthesis task.}
\end{itemize}
\section{Related Works}
\label{sec:related_works}

\renewcommand{\arraystretch}{1.0}
\begin{table}[t]
\small
\centering
\resizebox{0.48\textwidth}{!}{%
\begin{tabular}{lcccccccc}
\toprule
Datasets & NeRF\cite{mildenhall2020nerf} & T\&T\cite{knapitsch2017tanks} &  NeRF360\cite{barron2022mip}& BMVS\cite{yao2020blendedmvs} & MP3D\cite{Matterport3D} & DTU\cite{aanaes2016large}&\textbf{Ours}\\
\midrule
\# Scenes & 8 & 15 & 9 & 502 & 90 & 80 & 75 \\
\# Images & 0.8k & 2.8k & 2k & 101k & 190k& 3.9k & 15k \\
Multi-Object & \redxmark & \redxmark & \redxmark & \greencheckmark & \greencheckmark & \redxmark  & \greencheckmark \\
Outdoor scenes & \redxmark & \greencheckmark& \greencheckmark & \greencheckmark &  \redxmark & \redxmark  & \greencheckmark \\
Compositionality & \redxmark & \redxmark & \redxmark & \redxmark & \redxmark & \greencheckmark & \greencheckmark \\
360 Camera & \redxmark & \greencheckmark & \greencheckmark & \redxmark & \greencheckmark &  \redxmark & \greencheckmark          \\
All GT Annotations & \greencheckmark & \redxmark & \redxmark & \redxmark & \redxmark & \redxmark & \greencheckmark \\
\bottomrule
\end{tabular}
}
\caption{\textbf{Comparison of NeRDS 360 with prior novel view synthesis datasets} comparing \#~scenes, scale, whether scenes contain multi-objects, are outdoor~i.e. urban setting, supports compositionality, provides 360 cameras~(as opposed to front-facing), and all ground truth information is available~(\eg, depth, 3D bounding boxes, instance masks)}
\label{tab:dataset_tab}
\vspace{-5mm}
\end{table}
\noindent\textbf{Neural Implicit Representations} use neural networks to map euclidean or temporal coordinates to target scene properties~\cite{tewari2021advances}. These methods have been successfully used to represent 3D shapes with signed distances~\cite{park2019deepsdf,duan2020curriculum, irshad2022shapo, heppert2023carto, zakharov2022road} or occupancies~\cite{mescheder2019occupancy,takikawa2021nglod}. While earlier methods require ground truth 3D supervision, recent advances in differentiable neural rendering have enabled self-supervised learning of the target signal from only image supervision~\cite{sitzmann2019scene,mildenhall2020nerf}, with Neural Radiance Fields (NeRFs) achieving impressive results, particularly for novel view synthesis. NeRF extensions focus on reducing aliasing effects via multiscale representations~\cite{barron2021mip}, modeling unbounded scenes~\cite{martinbrualla2020nerfw}, disentangled object-background representations and blending~\cite{ost2021neural}, compositional generative models~\cite{Niemeyer2020GIRAFFE} and improving reconstruction and depth estimation accuracy via multi-view consistent features~\cite{stier2021vortx,sun2021neuralrecon,guizilini2022depth}.

\looseness=-1
\noindent\textbf{Generalizable and Feature-conditioned representations:} Neural implicit representations operate in the overfitting context, aiming to accurately encode a single scene/object~\cite{mescheder2019occupancy,mildenhall2020nerf,takikawa2021nglod,muller2022instant}. Other approaches use global conditioning (a single global latent vector) to learn object latent spaces~\cite{park2019deepsdf,jang2021codenerf,mueller2022autorf,rematas2021sharf, irshad2022shapo}, struggling to capture detailed high-frequency information. While usually limited to modeling single scenes, Pixel-NeRF~\cite{yu2020pixelnerf}, Scene Representation Transformers~\cite{srt22}, and GRF~\cite{trevithick2021grf} use local features as conditioners for generalizable neural fields. Recently, grounplanar~\cite{sharma2022seeing} and tri-planar representations gain popularity for efficiently representing scenes using hybrid implicit-explicit representation. EG3D~\cite{chan2022efficient} and GAUDI~\cite{bautista2022gaudi} utilize tri-planar representations in their generative models, employing adversarial training. Notably, GAUDI utilizes GAN inversion for conditional synthesis. Our approach, NeO 360,  does not rely on expensive GAN inversion and directly outputs density and RGB for new views in a feedforward manner.

\noindent\textbf{Novel View Synthesis Datasets:} Existing datasets for novel view synthesis fall under the following major categories:~1.~Synthetic scenes which hemispherical 360-degree views around an object of interest offering dense camera overlap for fine-grained reconstruction (these include SRN rendering~\cite{choy20163d}, RTMV~\cite{tremblay2022rtmv} and Google Scanned Objects~\cite{downs2022google}),~2.~Forward-facing scenes that move the camera in the vicinity of an object without providing full 360-degree coverage (these include LLFF~\cite{mildenhall2019llff}, DTU~\cite{aanaes2016large}, Blender MVS~\cite{yao2020blendedmvs}) and~3.~360-degree real-scenes which provide full surrounding coverage (these include Tanks and Templates~\cite{knapitsch2017tanks}, MipNeRF360 dataset~\cite{barron2021mip} and CO3D~\cite{reizenstein2021common}). These datasets mostly evaluate on indoor scenes and provide little or no compositionality~(i.e.~multi-objects, 3D bounding boxes) for training or evaluation. While MipNeRF360~\cite{barron2021mip} and Tanks and Tempalte~\cite{knapitsch2017tanks} provide 360 coverage, the number of scenes in these datasets is small; hence it is difficult evaluating the performance of generalizable NeRF methods at scale. Due to these challenges, we collect a large-scale outdoor dataset offering similar camera distributions as NeRF~\cite{mildenhall2020nerf} for 360$^{\circ}$ outdoor scenes. Our dataset, described in detail in~(Section~\ref{sec:dataset}), offers dense viewpoint annotations for outdoor scenes and is significantly larger than existing outdoor datasets for novel-view synthesis. This allows building effective priors for large-scale scenes which can lead to improved generalizable performance on new scenes with very limited views, as we show in Section~\ref{sec:experiments}.
%-------------------------------------------------------------------------

%-------------------------------------------------------------------------
\section{NeRDS 360 Multi-View Dataset for Outdoor Scenes}
\label{sec:dataset}
Due to the challenge of obtaining accurate ground-truth 3D and 2D information~(such as denser viewpoint annotations, 3D bounding boxes, and semantic and instance maps), only a handful of outdoor scenes have been available for training and testing.
Specifically, previous formulations~\cite{Ost_2021_CVPR, fu2022panoptic, kundu2022panoptic, rematas2022urban} have focused on reconstructions using existing outdoor scene datasets~\cite{geiger2012we, liao2022kitti, cabon2020vkitti2} offering panoramic-views from the camera mounted on an ego-vehicle. These datasets~\cite{nuscenes2019, sun2020scalability} offer little overlap between adjacent camera views~\cite{xie2023s}, a characteristic known to be useful for training NeRFs and multi-view reconstruction methods. Moreover, the optimization of object-based neural radiance models for these scenes becomes more challenging as the ego car is moving fast and the object of interest is observed in just a few views~(usually less than 5).

\begin{figure}[!t]
    \centering
    \resizebox{\linewidth}{!}{%
      \includegraphics{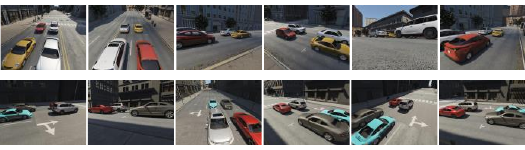}
    }
    \caption{\textbf{Proposed multi-view dataset} RGB renderings for 360$^{\circ}$ novel-view synthesis of outdoor scenes. 
    }
    \vspace{-2mm}
    \label{fig:dataset}
\end{figure}

\begin{figure}
    \centering
    \resizebox{\linewidth}{!}{%
      \includegraphics{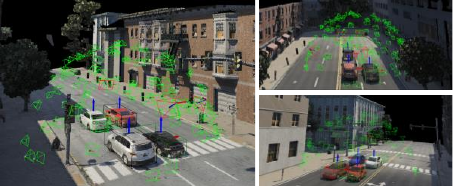}
    }
    \caption{Camera distribution for 1, 3 and 5 source views~(shown in \textcolor{red}{red}) and evaluation views~(shown in \textcolor{green}{green})
    }
    \vspace{-4mm}
    \label{fig:testcameras}
\end{figure}

\textbf{Dataset:} To address these challenges, we present a large-scale dataset for 3D urban scene understanding. Compared to existing datasets, as demonstrated in Table~\ref{tab:dataset_tab}, our dataset consists of 75 outdoor urban scenes with diverse backgrounds, encompassing over 15,000 images. These scenes offer 360$^{\circ}$ hemispherical views, capturing diverse foreground objects illuminated under various lighting conditions. Additionally, our dataset~(as shown in Fig.~\ref{fig:dataset},~\ref{fig:testcameras}) encompasses scenes that are not limited to forward driving views, addressing the limitations of previous datasets such as limited overlap and coverage between camera views~\cite{geiger2012we, liao2022kitti}). The closest pre-existing dataset for generalizable evaluation is DTU~\cite{aanaes2016large}~(80 scenes) which comprises mostly indoor objects and does not provide multiple foreground objects or background scenes.

We use the Parallel Domain~\cite{parallel_domain} synthetic data generation to render high-fidelity 360$^{\circ}$ scenes. We select 3 different maps i.e. \textit{SF 6thAndMission}, \textit{SF GrantAndCalifornia} and \textit{SF VanNessAveAndTurkSt} and sample 75 different scenes across all 3 maps as our backgrounds~(All 75 scenes across 3 maps are significantly different road scenes from each other, captured at different viewpoints in the city). We select 20 different cars in 50 different textures for training and randomly sample from 1 to 4 cars to render in a scene. We refer to this dataset as~\textbf{NeRDS 360}:~\textbf{Ne}RF for~\textbf{R}econstruction,~\textbf{D}ecomposition and Scene~\textbf{S}ynthesis of 360$^{\circ}$ outdoor scenes. In total, we generate 15k renderings~(Fig.~\ref{fig:dataset}) by sampling 200 cameras in a hemispherical dome at a fixed distance from the center of cars. We held out 5 scenes with 4 different cars and different backgrounds for testing, comprising 100 cameras distributed uniformly sampled in the upper hemisphere, different from the camera distributions used for training. We use the diverse validation camera distribution to test our approach's ability to generalize to unseen viewpoints as well as unseen scenes during training. As shown in Figure~\ref{fig:testcameras} and supplementary material, our dataset and the corresponding task is extremely challenging due to occlusions, diversity of backgrounds, and rendered objects with various lightning and shadows. Our task entails reconstructing 360$^{\circ}$ hemispherical views of complete scenes using a handful of observations i.e. 1 to 5 as shown by red cameras in Fig.~\ref{fig:testcameras} whereas evaluating using all 100 hemispherical views, shown as green cameras in Fig.~\ref{fig:testcameras}; hence our task requires strong priors for novel view synthesis of outdoor scenes. 

\section{Method}
\label{sec:method}
\begin{figure*}[t!]
\centering
\includegraphics[width=1.0\textwidth]{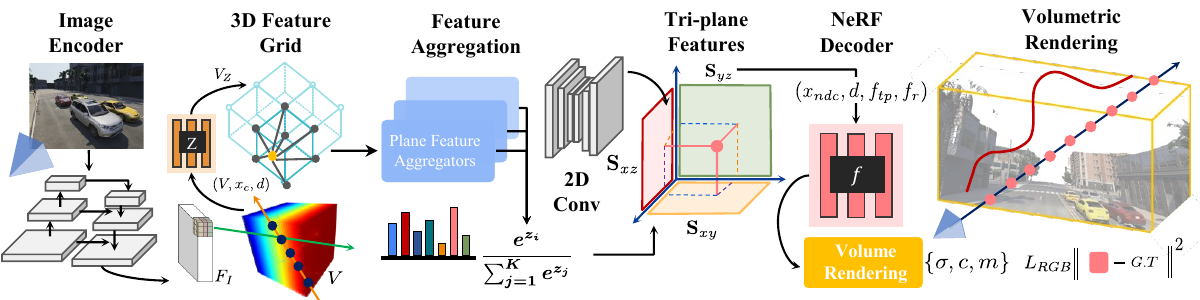}
\captionof{figure}{
\textbf{Method:} {Our method effectively uses local features to infer an image-conditional triplanar representation for both backgrounds and foregrounds. These triplanar features are obtained after orthogonally projecting positions~($x$) into each plane and bilinearly interpolating feature vectors. Dedicated NeRF decoder MLPs are used to regress density and color each for foreground and background.
}}
\label{fig:framework}
\vspace{-5mm}
\end{figure*}

\looseness=-1
Given RGB images of a few views of a novel scene, \titleShort{} infers a 3D scene representation capable of performing novel view synthesis and rendering 360$^{\circ}$ scenes. To achieve this goal, we employ a  hybrid local and global feature representation comprised of a triplanar representation that can be queried for any world point. Formally, as shown in Fig.~\ref{fig:teaser}, given a few input images,~$I=[I_{1}...I_{n}]$ of a complex scene, where $n =1$ to $5$, and their corresponding camera poses, $\gamma=[\gamma_{1}...\gamma_{n}]$ where $\gamma=[R|T]$, \titleShort{} 
infers the density and radiance fields for both near and far backgrounds~(similar to NeRF++~\cite{zhang2020nerf++}) with the major difference of using hybrid local and global features for conditioning the radiance field decoders instead of just positions and viewing directions, as employed in the classical NeRF formulation~\cite{mildenhall2020nerf, zhang2020nerf++}. We  describe our 3D scene representation in Section~\ref{sec:triplanar}, introduce deep residual local features in Section~\ref{sec:residual_features}, describe how we decode radiance fields conditioned on hybrid local and global features in Section~\ref{sec:decoding}, and discuss performing inference-time scene editing and composition in Section~\ref{sec:editing}.

\subsection{Image-Conditional Triplanar Representation}
\label{sec:triplanar}
\looseness=-1 \textbf{Preliminaries}: NeRF is an implicit 3D scene representation that learns a neural network $f(\mathbf{x}, \theta) \rightarrow(\mathbf{c}, \sigma)$. This end-to-end differentiable function $f$ outputs color $c_{i}$ and density $\sigma_{i}$ for every query 3D position $x_{i}$ and the viewing direction $\theta_{i}$ as input. For each point evaluation, a 4 channel~$(\mathbf{c}, \sigma)$ value is output, which is then alpha composited~(Eq. \ref{eq:alpha_comp} below) to render an image using volume rendering. 
\begin{equation}
\label{eq:alpha_comp}
\mathbf{c}=\sum_i w_i \mathbf{c}_i, \quad \mathbf{acc}=\sum_i w_i
\end{equation}
\begin{equation*}
w_i=\alpha_i \prod_{j<i}\left(1-\alpha_j\right), \quad \alpha_i=1-\exp \left(-\sigma_i\left\|\boldsymbol{x}_i-\boldsymbol{x}_{i+1}\right\|\right)
\end{equation*}

\textbf{Method:} Although producing high-fidelity scene synthesis, NeRF~\cite{mildenhall2020nerf} is limited in its ability to generalize to novel scenes. In order to effectively use scene-priors and learn from a large collection of unbounded 360$^{\circ}$ scenes, we propose an image-conditional triplanar representation, as shown in Fig.~\ref{fig:framework}. This representation is capable of modeling 3D scenes with full expressivity at scale without omitting any of its dimensions~(as in 2D or BEV-based representation) and avoiding cubic complexity~(as in voxel-based representations). Our triplanar representation comprises of three axis-aligned orthogonal planes~$S=[\textbf{S}_{xy}, \textbf{S}_{xz}, \textbf{S}_{yz}]$, $\in$  $\mathbb{R}^{3\times C \times D \times D}$ where $D \times D$ is the spatial resolution of each plane with feature $C$. 

To construct feature triplanes from input image, we first extract low-resolution spatial feature representations by using an  ImageNet~\cite{deng2009imagenet} pre-trained ConvNet backbone \textbf{E} which transforms input image $I \in \mathbb{R}^{H_i \times W_i \times 3}$ to 2D feature map $F_{I} \in \mathbb{R}^{H_i/2 \times W_i/2 \times C}$. Similar to prior works involving volumetric reconstruction~\cite{sun2021neuralrecon, murez2020atlas, kar2017learning}, the obtained local features are projected backwards along every ray to the 3D feature volume~($V_{F}$) using camera pose~$\gamma_{i}$ and intrinsic~$K_{i}$. 
While volumetric reconstruction methods~\cite{sun2021neuralrecon, murez2020atlas}, traditionally use the generated volume solely for indoor geometry reconstruction through TSDF, we show that it can also be employed in a computationally efficient way to estimate the entire scene's appearance and enable accurate neural rendering. Since all features along a camera ray are identical in the grid, we further learn depth of individual features by an additional MLP, $V_{Z} = Z(V_{F}, x_{c}, d)$ which takes as input concatenated features in the grid, positions of grid in the camera frame~($x_{c}$) and directions from the positions of grid in the world frame~$x_{w}$ to the camera frame and outputs depth-encoded features $V_{Z}$. Next, we obtain triplane features using learnt weights~($w_{i}$) over individual volumetric feature dimensions:
\begin{align}
\textbf{S}_{xy} =\sum_i w_{xy_i} \mathbf{V}_{Z_i},  \quad w_{xy} &= A_{xy}(V_{Z_i}, x_{w_z}) \\
\textbf{S}_{xz} =\sum_i w_{xz_i} \mathbf{V}_{Z_i},\quad   w_{xz} &= A_{xz}(V_{Z_j}, x_{w_y}) \\
\textbf{S}_{yz} =\sum_i w_{yz_i} \mathbf{V}_{Z_i},\quad w_{yz} &= A_{yz}(V_{Z_j}, x_{w_x})
\end{align}

where $A_{xy}, A_{xz}$ and $A_{yz}$ denote feature aggregation MLPs and $w_{xy}, w_{xz}$ and $w_{yz}$ are softmax scores obtained after summing over the $z, y$ and $x$ dimensions respectively. One motivation to project features into respective planes is to avoid the computationally cubic complexity of 3D CNNs as in~\cite{chen2021mvsnerf, stier2021vortx} and at the same time be more expressive than BEV or 2D feature representations~\cite{li2022bevdepth, li2022bevformer, yu2020pixelnerf} which are computationally more efficient than voxel-based representations but omitting z-axis hurts their expressiveness. We instead rely on 2D convolutions to transform the built image-conditional triplanes into a new G-channel output, where $G=C/4$, while upsampling the spatial dimension of planes from $D\times D$ to image feature space (i.e. $H/2 \times W/2$). The learnt convolutions act as inpainting networks to fill in missing features.  
As shown in Fig.~\ref{fig:framework}, our triplanar representation acts as a global feature representation,  as intuitively a complex scene can be better represented when examined from various perspectives. This is because each may offer complementary information that can help understand the scene more effectively.

\subsection{Deep Residual Local Features:}
\label{sec:residual_features}
As noted by~\cite{chen2021mvsnerf, wang2021ibrnet} and inspired by ~\cite{heDeepResidualLearning2016}, for the following radiance field decoding stage, we also use the features $f_{r}$ as a residual connection into the rendering MLP. We obtain $f_{r}$ from $F_{I}$ by projecting the world point $x$ into source view using its camera parameters $\gamma_{i},K_{i}$ and extracting features at the projected pixel locations through bilinear interpolation similar to~\cite{yu2020pixelnerf}. Note that both local and global feature extraction pathways share the same weights~$\theta_{E}$ and encoder~$\textbf{E}$. We find that for complex urban unbounded scenes, using just local features similar to~\cite{yu2020pixelnerf} leads to ineffective performance for occlusions and faraway 360$^{\circ}$ views, as we show quantitatively and qualitatively in Section~\ref{sec:experiments}. Using only global features, on the other hand, leads to hallucinations, as shown in our ablations~\ref{sec:experiments}. Our method combines both local and global feature representations effectively, resulting in a more accurate 360$^{\circ}$ view synthesis from as minimal as a single view of an unbounded scene.
  
\subsection{Decoding Radiance Fields:}
\label{sec:decoding}
The radiance field decoder $D$ is tasked with predicting color $c$ and density $\sigma$ for any arbitrary 3D location $x$ and viewing direction~$d$ from triplanes $S$ and residual features $f_{r}$. We use a modular implementation of rendering MLPs, as proposed by ~\cite{zhang2020nerf++} with the major difference of using our local and global features for conditioning instead of just using positions and viewing diretions as an input to the MLPs. The MLP is denoted as: 
\begin{equation}
\label{eq:rendering}
\sigma, c = D(x, d, f_{tp}, f_{r})    
\end{equation}
where we obtain $f_{tp}$ by orthogonally projecting point $x$ into each plane in $S$ and performing bi-linear sampling. We concatenate the three bi-linearly sampled vectors into $f_{tp} = [\textbf{S}_{xy}(i, j),\textbf{S}_{xz}(j, k),\textbf{S}_{yz}(i, k)]$. Note that similar to ~\cite{yu2020pixelnerf}, we establish our coordinate system using the view space of the input image, and then indicate the positions and camera rays within this particular coordinate system. By utilizing the view space, our method can successfully standardize the scales of scenes from various data sources, thereby enhancing its ability to generalize well. Although our method gives reasonable results from single-view observation~(Section~\ref{sec:experiments}), \titleShort{} can seamlessly integrate multi-view observations by pooling along the view dimension in the rendering MLPs. We refer to our supplementary material for a detailed architecture diagram and description. 

\textbf{Near and Far Decoding MLPs:}
Similar to NeRF++~\cite{zhang2020nerf++}, we define two rendering MLPs for decoding color and density information as follows:
\begin{equation}
D(.) =\begin{cases}
D_{fg}(.) & \text{if } \sum\limits_{i=1}^n (x_i^2+y_i^2+z_i^2) < 1 \\
D_{bg}(.) & \text{if } \sum\limits_{i=1}^n (x_i^2+y_i^2+z_i^2) > 1
\end{cases}
\end{equation}
where we define a coordinate remapping function~($M$) similar to the original NeRF++ formulation~\cite{zhang2020nerf++} to contract the 3D points that lie outside the unit sphere where $M$ maps points~($x,y,z$) outside the unit sphere to the new 4D coordinates$
\left(x^{\prime}, y^{\prime}, z^{\prime}, 1 / r\right)
$, where $(x^{\prime}, y^{\prime}, z^{\prime})$ represents the unit vector in the direction of ($x,y,z$) and $r$ denotes the inverse radius along this dimension. This formulation helps further objects get less resolution in the rendering MLPs. For querying our tri-planar representation, we use the un-contracted coordinates ($x,y,z$) in the actual world coordinates since our representation is planes instead of spheres. For rendering, we use the respective contracted coordinates~($x^{\prime}, y^{\prime}, z^{\prime}$) for conditioning the MLPs.  
\begin{table*}
\centering
\renewcommand{\arraystretch}{0.95}
\footnotesize
% \begin{footnotesize}
% \begin{tabular}{ccrclrclrclrcrrcr}
\resizebox{\textwidth}{!}{%
\begin{tabular}{cccccccccccccc}
\toprule
\multirow{3}{*}{Method}  & \multirow{3}{*}{\# Views}   & \multicolumn{4}{c}{Single Map (Prior Sampling)} &\multicolumn{4}{c}{Single Map (Novel Scenes)}&\multicolumn{4}{c}{Multi-Map (Novel Scenes)}\\
\cmidrule(lr){3-6}\cmidrule(lr){7-10} \cmidrule(lr){11-14}

& & \multicolumn{3}{c}{\textbf{Scenes}} & \textbf{Objects} &\multicolumn{3}{c}{\textbf{Scenes }} & \textbf{Objects} &\multicolumn{3}{c}{\textbf{Scenes}} & \textbf{Objects}\\
\cmidrule(lr){3-5} \cmidrule(lr){6-6} \cmidrule(lr){7-9} \cmidrule(lr){10-10}
\cmidrule(lr){11-13} \cmidrule(lr){14-14}
& & \textbf{PSNR} $\uparrow$ & \textbf{SSIM} $\uparrow$ & \textbf{LPIPS} $\downarrow$ & \textbf{PSNR} $\uparrow$ & \textbf{PSNR} $\uparrow$ & \textbf{SSIM} $\uparrow$ & \textbf{LPIPS} $\downarrow$ & \textbf{PSNR} $\uparrow$& \textbf{PSNR} $\uparrow$ & \textbf{SSIM} $\uparrow$ & \textbf{LPIPS} $\downarrow$ & \textbf{PSNR} $\uparrow$\\ 

\midrule

\multirow{3}{1.8cm}{\textcolor{pink}{\Large{$\blacktriangle$}} \centering NeRF~\cite{mildenhall2020nerf}}
& 1 & 12.41 & 0.09 & 0.71 & 10.55 & 12.82 & 0.13 & 0.69 & 10.45 & 12.72& 0.12 & 0.69 & 10.39  \\
& 3 & 14.14 & 0.25 & 0.59& 11.67 & 13.76 & 0.18 & 0.62 & 11.81 & 13.82& 0.19& 0.61&  11.85\\
& 5 & 15.37 & 0.38 & 0.50& 12.89 & 16.16 & 0.39 & 0.49 & 14.83 & 16.14& 0.38& 0.48& 14.73\\
\midrule
\multirow{3}{1.8cm}{\textcolor{pink}{\Large{$\blacktriangle$}} \centering MVSNeRF~\cite{chen2021mvsnerf}}
& 1 &  14.40 & 0.40 & 0.65 & 12.42& 13.87 & 0.33 & 0.65 & 11.08 & 13.90&  0.31&  0.65& 11.10\\
& 3 & 13.93 & 0.34 & 0.63 & 11.33 & 14.50 & 0.40 & 0.64 & 12.79 &14.40 & 0.38&  0.63& 12.70\\
& 5 & 14.78 & 0.39 & 0.62 & 12.21& 15.43 & 0.41 & 0.61 & 14.13 & 15.40 & 0.42& 0.62&14.10\\
\midrule

\multirow{3}{1.8cm}{\textcolor{pink}{\Large{$\blacktriangle$}} \centering Pixel-NeRF~\cite{yu2020pixelnerf}}
& 1 & 15.89 & 0.44 & 0.64 & 13.57 & 14.93 & 0.40 & 0.65 & 12.93 & 15.01& 0.47 & 0.65&12.65\\
& 3 & 17.15 & 0.50 & 0.62 & 14.47 & 17.46 & 0.48 & 0.63 & 15.70 & 16.20 & 0.52 & 0.64& 13.00\\\
& 5 & 17.50 & 0.51 & 0.62 & 14.75 & 17.80 & 0.49 & 0.62 & 15.92 & 16.91 & 0.52& 0.62& 14.22\\
\midrule

\multirow{3}{1.8cm}{\textcolor{orange}{\Large{$\blacktriangle$}} \centering Pixel-NeRF\textsubscript{\textit{ft}} ~\cite{yu2020pixelnerf}}
& 1 & -& - & - & -& 15.78 & 0.43 & 0.66 & 14.07 & 14.65& 0.42 & 0.66&11.81 \\
& 3 & - & - &-& -& 17.90 & 0.51 & 0.59 & 17.12 & 16.69 & 0.54 & 0.62& 13.49\\
& 5 & - & - & - & - & 19.26 & 0.55 & 0.57 & 18.54& 17.22 & 0.55& 0.61& 15.21 \\
\midrule

\multirow{3}{1.8cm}{\textcolor{pink}{\Large{$\blacktriangle$}} \centering \textbf{\titleShort{}} (Ours)}
&1 & 16.91 & 0.51 & 0.56 & 14.11 & 17.60 & 0.56& 0.51&15.80 & 16.30 &0.52 & 0.57& 13.04\\
& 3 & 18.94 & 0.58 & 0.48 & 16.66 & 19.35 & 0.59& 0.50 & 17.60 & 18.59&0.61 &0.52 & 15.93\\

& 5 & \underline{19.64} & \underline{0.62} & \underline{0.47} & \underline{17.34} & 20.10 & 0.62& 0.48& 18.20 & 19.27& 0.64 & 0.49& 16.60\\
\midrule
\multirow{3}{1.8cm}{\textcolor{orange}{\Large{$\blacktriangle$}} \centering 
 \textbf{\titleShort{}}\textsubscript{\textit{ft}} (Ours)}
&1 & - & - & - & - &{17.93} & {0.58}& {0.49}&{15.95} & 16.42& 0.55& 0.54& 13.80 \\
& 3 & - & - & - & - & 19.56 & {0.61} & 0.46& 18.30 & 18.94 & 0.63 & 0.49& 16.81\\
& 5 & - & -& -& - & \underline{20.56} & \underline{0.64}& \underline{0.45}& \underline{18.62} &\textbf{19.59}&\textbf{0.67} &\textbf{0.46}&\textbf{17.70}\\
\bottomrule
\end{tabular}
}
% \end{footnotesize}
\caption{\small{\textbf{Quantitative novel view synthesis results}: Conditional prior-based sampling and novel-scene rendering. \textcolor{pink}{Pink} \textcolor{pink} {\Large{$\blacktriangle$}}  denotes zero-shot evaluation whereas \textcolor{orange}{Orange}\textcolor{orange} {\Large{$\blacktriangle$}} denotes finetuning only the triplanar network i.e. freezing the encoder \textbf{E} with learning rate $5 \times 10^{-6}$
from 1,3 or 5 source views. The orange triangle for PixelNeRF denotes that we finetune their encoder network similar to how we finetune our triplanes. \underline{Underline} shows best results when trained and evaluated within the same single map~(easier setting), \textbf{Bold} denotes best results with challenging evaluation setting where the evaluation dataset is sampled across 3 different maps with diverse illumination.}
}
\label{table:overfitting_objects}
\vspace{-2mm}
\end{table*}

\textbf{Optimizing radiance fields for few-shot novel-view synthesis:}
Given local and global features constructed from source views, we decode color $c^{i}_{p}$ and density $\sigma^{i}_{p}$ for backgrounds using dedicated near and far background MLPs $D_{near}(.)$ and $D_{far}(.)$~(Eq.~\ref{eq:rendering}) after volumetrically rendering and compositing the near and far backgrounds and enforcing the loss as follows:
\begin{equation}
    L=\left\|c_p-\tilde{c}_t\right\|_2^2 +\lambda_{reg} L_{reg} + \lambda_{LPIPS} L_{LPIPS}
\end{equation}

where $\tilde{c}_t$ is the sampled pixel locations from the target image and $c^{p}$ is the composited color obtained from the rendering output of near and far MLPs as~$c^{p}_{i} =  c^{nb}_{i} + \prod_{j<i}\left(1-\alpha^{nb}_j\right) c^{b}_{i}$. We also encourage the weights of near and far background MLPs to be sparse for efficient rendering by enforcing an additional distortion regularization loss~\cite{barron2021mip} and further use ${L}_{LPIPS}$ loss to encourage perceptual similarity b/w patches of rendered color, $c^{p}$ and ground color~$\tilde{c}_t$, where we only enforce it after 30 training epochs to improve background modeling ~(c.f. see supplementary). 

\subsection{Scene Editing and Decomposition:}
\label{sec:editing}

\begin{figure*}[ht!]
\centering
\includegraphics[width=1.0\textwidth]{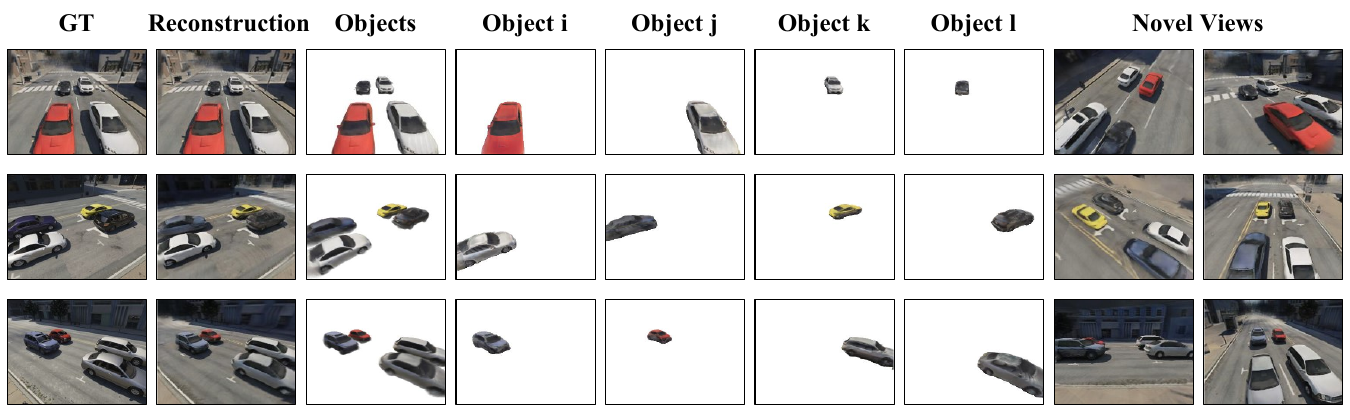}
\captionof{figure}{
\textbf{Scene decomposition qualitative results} showing 3-view scene decomposed individual objects along with novel views on the NeRDS 360 dataset. Our approach performs accurate decomposition by sampling inside the 3D bounding boxes of the objects; hence giving full control over object editability from very few input views.
}
\vspace{-8pt}
\label{fig:qualitative_main}
\end{figure*}

Given 3D bounding boxes obtained from a detector, we can obtain the individual object and background radiance fields for each object by simply sampling rays inside the 3D bounding boxes of the objects and bilinearly interpolating the features at those specific~($x,y,z$) locations in our tri-planar feature grid (\textbf{S}), making it straightforward to edit out and re-render individual objects. As
illustrated in Fig.~\ref{fig:qualitative_main}, we perform accurate object re-rendering by considering the features inside the 3D bounding boxes of objects to render the foreground MLP. In essence, we divide the combined editable scene rendering formulation as rendering objects, near backgrounds, and far backgrounds. For far backgrounds, we retrieve the scene color~$c^{b}_{i}$ and density~$\sigma^{b}_{i}$ which is unchanged from the original rendering formulation. For near backgrounds, we obtain color~$c^{nb}_{i}$ and density~$\sigma^{nb}_{i}$ after pruning rays inside the 3D bounding boxes of objects~(i.e. setting $\sigma^{nb}_{i}$ to a negative high value, $-1 \times 10^{-5}$ before volumetrically rendering). For objects, we only consider rays inside the bounding boxes of each object and sampling inside foreground MLP to retrive~$c^{o}_{i}$ and density~$\sigma^{o}_{i}$. We aggregate the individual opacities and colors along the ray to render composited color using the following equation:
\begin{equation}
\label{eq:alpha_comp_combined}
\mathbf{c}=\sum_i w^{b}_i \mathbf{c}^{b}_i + \sum_i w^{nb}_i \mathbf{c}^{nb}_i + \sum_i w^{o}_i\mathbf{c}^{o}_i
\end{equation}
\section{Experiments}
\label{sec:experiments}
\looseness=-1
We evaluate our proposed method against various state-of-the-art baselines, focusing on few-shot novel view synthesis including~\textbf{a.} Conditional prior-based sampling and~\textbf{b.} Novel scene rendering tasks. We compare full scenes on the following baselines: 1) \textbf{NeRF}~\cite{mildenhall2020nerf}: Vanilla NeRF formulation which overfits to a scene given posed RGB images 2) \textbf{PixelNeRF}~\cite{yu2020pixelnerf} A generalizable NeRF variant which utilizes local image features for few-shot novel view synthesis 3) \textbf{MVSNeRF}~\cite{chen2021mvsnerf}: Extends NeRF for few-view synthesis using local features obtained by building a cost-volume from source images and 4)~\textbf{\titleShort{}}: Our proposed architecture which combines local and global features for generalizable scene representation learning. 

\textbf{Metrics:}
We use standard PSNR, SSIM, and LPIPS metrics to evaluate novel-view synthesis and L1 and RMSE to measure depth reconstruction quality.
\renewcommand{\arraystretch}{0.8}
\begin{table}[t]
\centering
\resizebox{0.40\textwidth}{!}{%
\begin{tabular}{@{}lccc@{}}
\toprule
\multirow{2}{*}{Method}& \multicolumn{2}{c}{\textbf{Backgrounds}} & \textbf{Objects} \\
\cmidrule(lr){2-3} \cmidrule(lr){4-4}
 & \textbf{PSNR} $\!\uparrow$ & \textbf{SSIM} $\!\uparrow$ & \textbf{PSNR} $\!\uparrow$ \\ \midrule
NeRF~\cite{mildenhall2019llff} (No Priors) & 16.16 & 0.34 & 15.42\\
Ours (No Pretraining) & 16.40 & 0.39 & 15.70\\
\midrule
Ours & \textbf{20.48} & \textbf{0.67} & \textbf{19.03}\\ 
\bottomrule
\end{tabular}
}
\caption{\label{tab:object_discovery} \small{Effect of scene priors on 3-view novel view synthesis on NeRDS 360 dataset~(Single Map)}}
\vspace{-6mm}
\end{table}

\textbf{Comparison with strong baselines for novel-view synthesis:} We aim to answer the following key questions: \textbf{1.} Does our generalizable tri-planar representation perform better than other generalizable NeRF variants given access to prior data and a few views for optimization on novel scenes? \textbf{2.} Do scene priors help with zero-shot generalization? and \textbf{3.} Does scaling the data help our network generalize better? We summarize the results in Table~\ref{table:overfitting_objects} and note that \titleShort{} achieves superior results compared to state-of-the-art generalizable NeRF variants i.e. PixelNeRF~\cite{yu2020pixelnerf} and MVSNeRF~\cite{chen2021mvsnerf} in both zero-shot testing and finetuning given a limited number of source views. Spefically~\titleShort{} achieves a PSNR of 19.35, SSIM of 0.59, and LPIPS of 0.50 for complete scenes and 17.60 PSNR for objects with zero-shot evaluation, hence demonstrating an absolute improvement of 1.89 PSNR, 0.11 SSIM, 0.13 LPIPS for complete scenes and 1.90 PSNR for novel objects against the best-performing baseline for a single-map scenario~(i.e. all methods trained on 25 scenes and evaluated on a single novel scene within the same map). Our approach also outperforms the best baselines on the challenging multi-map dataset with zero-shot evaluation, where 5 novel scenes were held out with difficult illuminations and shadows.~\titleShort{} achieves an absolute PSNR improvement of 2.39 on complete scenes and 2.93 on objects, showing that~\titleShort{} learns better priors for unbounded scenes. The table also shows both NeO 360 and PixelNeRF perform better than the original NeRF since NeRF has not seen any scene level prior and is optimized per scene from only a few images.

\renewcommand{\arraystretch}{0.8}
\begin{table}[t]
\centering
\resizebox{0.40\textwidth}{!}{%
\begin{tabular}{@{}lccc@{}}
\toprule
 Method & \textbf{PSNR} $\!\uparrow$ & \textbf{SSIM}
 $\!\uparrow$ & \textbf{LPIPS} $\!\downarrow$ \\
\cmidrule(r){1-1} \cmidrule(lr){2-4}
with Colors & 16.29 & 0.44 & 0.62\\
w/o Feature Grid  & 17.50 & 0.47 & 0.59\\
w/o Near/Far & 19.02 & 0.57  & 0.52  \\
\midrule
Ours (global + local) & \textbf{19.35} & \textbf{0.59} & \textbf{0.50}\\ 
\bottomrule
\end{tabular}
}
\caption{\label{tab:ablation_architecture} \small{Effect of different design choices in our architecture for 3-view novel view synthesis on NeRDS 360 dataset~(Single Map)}}
\vspace{-6mm}
\end{table}
\textbf{Ablation Analysis:} We further show the effect of our design choices in Table~\ref{tab:ablation_architecture}. We show that directly using colors from source views results in a much worse performance of our model due to over-reliance on source view pixels which hurts generalization to out-of-distribution camera views.  Additionally, we ablate the feature grid as well as the near/far MLPs. The results confirm that performance degrades without the use of the 3D feature grid i.e. global features~(row 2) which is one of our major contributions. The near/far decomposition has a less significant but still positive effect. Overall, our model with combined local and global features in the form of triplanes performs the best among all variants.

\textbf{Effect of Scene Priors:}
Table~\ref{table:overfitting_objects} shows \titleShort{}'s ability to overfit to a large number of scenes, hence achieving better PSNR than PixelNeRF on sampling from prior distributions for novel trajectories. To further validate the effect of scene priors, we omit scene priors from our approach's training and the results are summarized in Table~\ref{tab:object_discovery}. The results further confirm that scene priors actually help our network, resulting in an absolute PSNR improvement of 4.08 comparing the network learned on 25 scenes with the network which has seen no prior scenes during training and is only overfit on 3 unseen views from scratch for a novel scene.  The results show that  the architecture design of our network allows us to learn from a large collection of scenes while extending the learned prior to novel scenes with effective zero-shot generalizability from a few views.

\begin{table}[t]
% \vspace{-2pt}
\begin{minipage}{.48\columnwidth}
\centering
\resizebox{0.99\columnwidth}{!}{
\begin{tabular}{@{}lcc@{}}
\toprule
Comparison       & \textbf{PSNR} ↑ & \textbf{SSIM} ↑  \\ 
\midrule
mipNeRF360~\cite{barron2022mip}      & 13.25 & 0.31  \\
SRT~\cite{srt22} & 14.61 & 0.40\\
EG3D*/GAUDI*~\cite{chan2022efficient}~\cite{bautista2022gaudi}   & 12.84 & 0.30 \\

\midrule
\textbf{NeO 360 (Ours)}   & \textbf{19.35} & \textbf{0.59}\\
\bottomrule
\end{tabular}
}
% \vspace{-6pt}
\captionof{table}{Quantitative comparison of 3-view view synthesis on NeRDS 360}
\label{tab:add_comparison}
\end{minipage}
\centering
\hfill\begin{minipage}{.48\columnwidth}
\centering
\resizebox{0.99\columnwidth}{!}{
\begin{tabular}{@{}lcc@{}}
\toprule
Object Depth & \textbf{L1} ↓ & \textbf{RMSE} ↓  \\ 
\midrule
PixelNeRF~\cite{yu2021pixelnerf} & 0.83 & 1.07  \\
NeO 360~(no ft.) & 0.59 & 0.74\\
\midrule
\textbf{NeO 360~(Ours)}   & \textbf{0.20} & \textbf{0.61}\\
\bottomrule
\end{tabular}

}
% \vspace{-5pt}
\captionof{table}{Eval views depth prediction from 3 source views on NeRDS 360}
\label{tab:depth_metrics}

\end{minipage}
\vspace{-16pt}
\end{table}

\begin{figure}[!b]
    \centering
    \resizebox{1.0\linewidth}{!}{%
\includegraphics{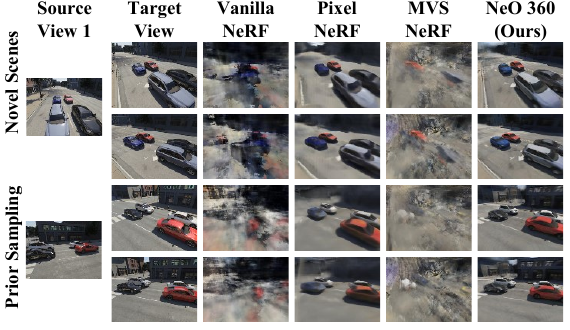}
    }
    \caption{\textbf{Qualitative 3-view view synthesis results}: Comparisons with baselines.
    }
    \label{fig:qualitative_all_baselines}
\end{figure}%
\textbf{Additional baseline comparisons:} We include additional comparisons with novel-view synthesis baselines~(mipNeRF360~\cite{barron2022mip}, EG3D~\cite{chan2022efficient} and SRT~\cite{srt22}) in Tab.~\ref{tab:add_comparison}. One could clearly observe that naively using triplanes~(Tab.~\ref{tab:add_comparison} row 3, *denotes that we take the triplanar representation without generative losses or training) or local features as in PixelNeRF~(Tab.~\ref{tab:ablation_architecture})  hurt the performance. Our method relies on effective combination of local and global features which serves as a strong baseline for the challenging task of 360$^{\circ}$ view synthesis of outdoor scenes. We also include depth reconstruction metrics~(Tab.~\ref{tab:depth_metrics}) and show our techniques' superior results compared to Pixel-NeRF.

\section{Qualitative Results:}
\textbf{Comparison with generalizable NeRF baselines:} As seen in Figure~\ref{fig:qualitative_all_baselines}, our method excels in novel-view synthesis from 3 source views, outperforming strong generalizable NeRF baselines. Vanilla NeRF struggles due to overfitting on these 3 views. MVSNeRF, although generalizable, is limited to nearby views as stated in the original paper, and thus struggles with distant views in this more challenging task whereas PixelNeRF's renderings also produce artifacts for far backgrounds. 

\begin{figure}[!b]
    \centering
    \resizebox{0.96\linewidth}{!}{%
\includegraphics{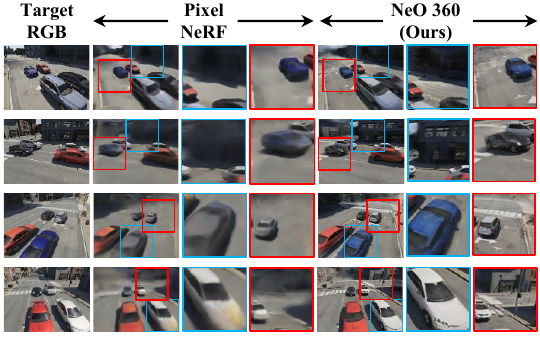}
    }
    \caption{\textbf{Qualitative 3-view view synthesis results}: Close-up comparison with PixelNeRF~\cite{yu2020pixelnerf}.
    }
    \label{fig:qualitative_pixelnerf}
\end{figure}%

\begin{figure*}[t]
\centering
\includegraphics[width=1.0\textwidth]{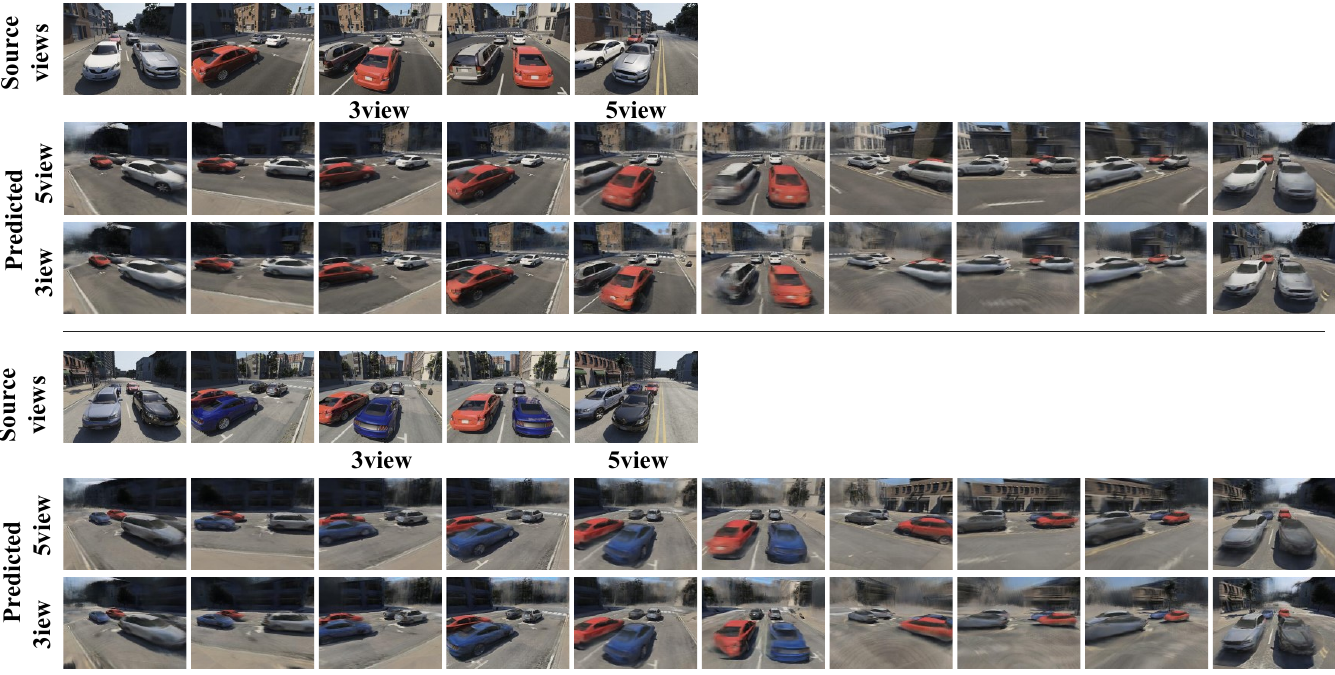}
\captionof{figure}{
\textbf{~\titleShort{}'s zero-shot qualitative results:} {We show 360$^{\circ}$ predictions for 3-view and 5-view novel view synthesis. Note that although our network has some shape artifacts for 3-view novel-view synthesis, these are effectively resolved by adding a few more sparse views, showing our network's ability to effectively use learned priors for sparse novel-view synthesis in a zero-shot manner. We show 10 predicted samples with indices 11, 20, 32, 38, 43, 48, 65, 76, 84, and 98 rendered from a circular trajectory generated at a consistent radius around the scene.
}}
\vspace{-5pt}
\label{fig:360_qualitative}
\end{figure*}

\textbf{Detailed comparison with PixelNeRF:} Figure~\ref{fig:qualitative_pixelnerf} presents our method's novel-view synthesis results compared with PixelNeRF. The red and blue boxes focus on close-ups. The visual comparison emphasizes our method's ability to produce crisper and clearer object and background renderings, while PixelNeRF generates blurrier outputs with noticeable artifacts in both foreground and background. 

\textbf{Scene Decomposition:} We further show our network's scene decomposition performance in Figure~\ref{fig:qualitative_main}. The figure demonstrates precise object recovery from the near background MLP output through sampling within each object's GT 3D bounding box (as emphasized in Section~\ref{sec:editing}).  This formulation allows us to easily re-render objects thanks to our feature-based representation which can be queried individually for objects and backgrounds. Note that we do not enforce any objectness prior during training to get this behavior, it is purely learned from multi-view image-based rendering. 

\textbf{360$^{\circ}$ qualitative results:} We further show our network's predicted few-view 360$^{\circ}$ novel view synthesis output in a zero-shot manner on unseen scenes and objects, not observed during training. As shown by Fig.~\ref{fig:360_qualitative}, our method performs plausible novel-view synthesis of complete scenes including far-away backgrounds and objects from very few sparse views of an outdoor scene; hence demonstrating~\titleShort{}'s ability to use learned priors effectively. We also show that 3-view synthesis introduces some artifacts in parts of the scene where there is no overlap between source views~i.e. where the scene is entirely unobserved. Fig.~\ref{fig:360_qualitative} shows that by adding a few sparse sets of views in those areas~(i.e.~5-view case), those artifacts can be effectively removed and a smooth scene representation could be obtained. This shows our network's ability to interpolate smoothly across given source views and also complete the scene in an effective manner.
\vspace{-15pt}
\section{Conclusion}
In this paper, we proposed \titleShort{}, a generalizable extension to the NeRF approach for unbounded 360$^{\circ}$ scenes. Our method relies on image-conditional tri-planar representations for few-shot novel view synthesis. In order to build strong priors for unbounded scenes, we propose a large-scale dataset, NERDS 360 to study view synthesis, reconstruction and decomposition in a 360-degree setting. Our method performs significantly better than other generalizable NeRF variants and achieves higher performance when tested on novel scenes. For future work, we will explore how the proposed method can be used to build priors that rely less on labelled data, such as 3D bounding boxes during inference and instead rely on motion cues for effective scene decomposition without labelled data. A second avenue of future work consists of sim2real extensions of this work to alleviate the data and annotations requirement in the real world by using only labelled data in simulation.

%------------------------------------------------------------------------

%%%%%%%%% REFERENCES
{\small
\bibliographystyle{ieee_fullname}
\bibliography{finalpaper}
}
\clearpage
\cleardoublepage
\begin{strip}
\begin{center}
\vspace{-5ex}
\textbf{\Large \bf
% Panoptic Segmentation in the Bird's Eye View
% Learning and Aggregating Lane Graphs for Urban Automated Driving
Supplementary Material for NeO 360: Neural Fields for \\
Sparse View Synthesis of Outdoor Scenes
} \\
\vspace{0.4cm}
\large{Muhammad Zubair Irshad$^1$$^2$\hspace{.2cm} Sergey Zakharov$^2$\hspace{.2cm} Katherine Liu$^2$\hspace{.2cm} Vitor Guizilini$^2$\hspace{.2cm} Thomas Kollar$^2$\\ 
Adrien Gaidon$^2$\hspace{.2cm} Zsolt Kira*$^1$\hspace{.2cm} Rares Ambrus*$^2$\\[0.01cm]
\footnotesize{* denotes shared last authorship}\\[0.09cm]}
\normalsize{$^1$Georgia Institute of Technology\hspace{.4cm} $^2$Toyota Research Institute\\[0.01cm]
}
\vspace{1.3cm}
\centering
\includegraphics[width=1.0\textwidth]{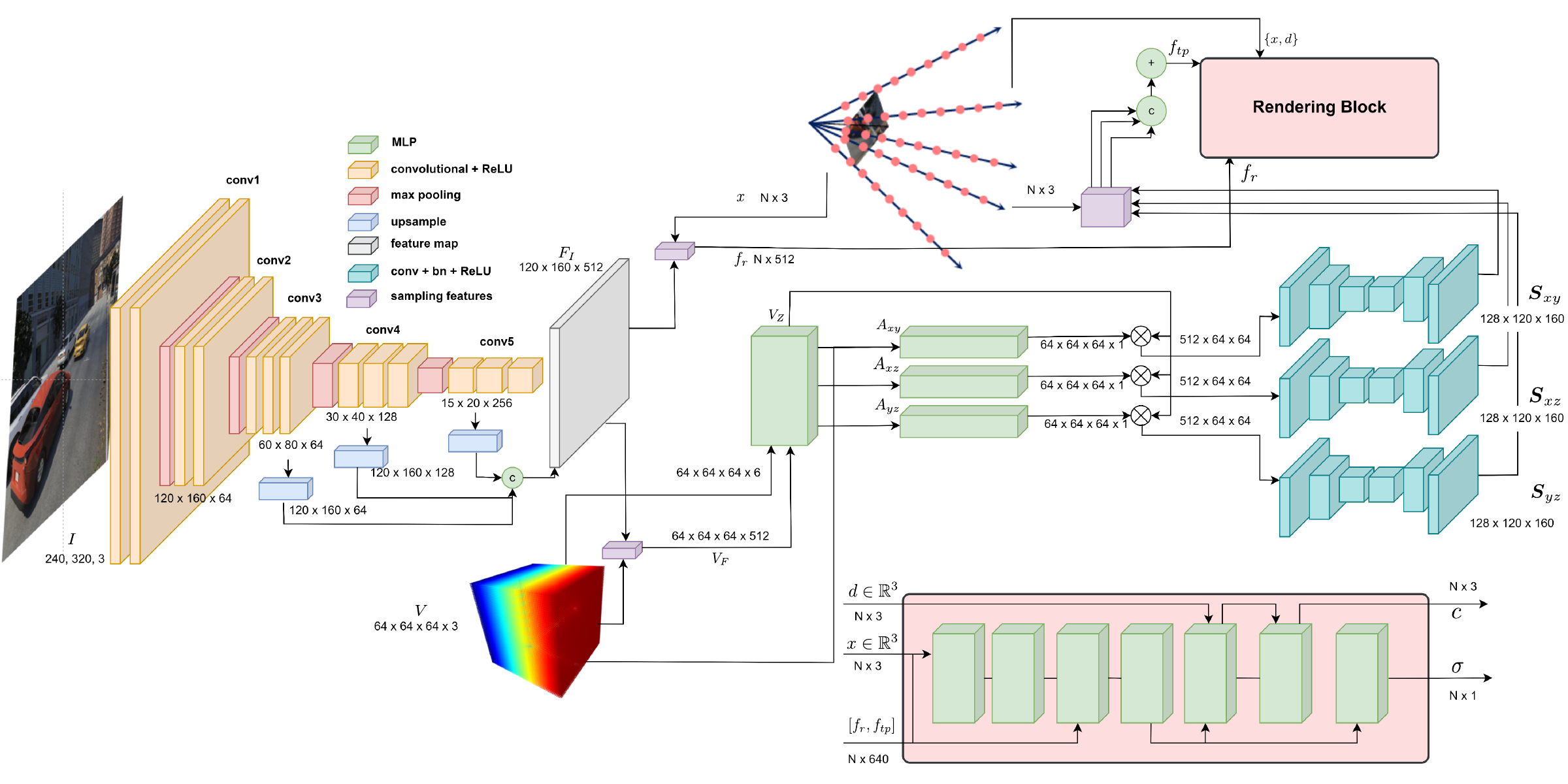}
    \captionsetup{width=\linewidth}
    \captionof{figure}{
    \textbf{Detailed Architecture}~ 
 of~\titleShort{} showing the construction of image conditional triplanes, along with local residual features and rendering MLPs to output density and color for a 3D point~$x$ and viewing direction~$d$.
    }   \label{fig:supp_architecture}
\end{center}
\end{strip}

\appendix
\setcounter{section}{0}
\setcounter{table}{0}
\makeatletter
%%%%%%%%% BODY TEXT
\normalsize
%%%%%%%%% ABSTRACT
\section{Network Architecture Details:}

\label{sec:architecture}
\looseness=-1
In this section, we provide more details about our architectural design, specifically our image-conditional tri-planar representation and our rendering MLPs as shown in Fig.~\ref{fig:framework} in the main paper. Our detailed architecture is presented in Fig.~\ref{fig:supp_architecture} and described in the following sections. We first describe the details of our encoder in Sec.~\ref{sec:encoder}, next we describe the details of our tri-planar features and residual features in Sec.~\ref{sec:features}. Finally we provide details of our rendering MLPs in Sec.~\ref{sec:rendering_mlps}.
\subsection{Encoder:}
\label{sec:encoder}
The Encoder network~$E$ comprises a pre-trained Resnet32~\cite{heDeepResidualLearning2016} backbone. We extract features from the initial convolutional layer and subsequent layer 1 to layer 3 and upsample all features to the same spatial resolution~(i.e. $H/2 \times W/2$) where $W,H$ are image dimensions respectively, before concatenating along the feature dimension to output feature map $F_{I}$ with dimension $512 \times H/2 \times W/2$ as shown in Fig.~\ref{fig:supp_architecture}.

\subsection{Image-conditional tri-planar features and residual features:}

In this section, we delve into the image-conditional tri-planar features and residual features. These elements, formed through volumetric projection, depth encoding, feature aggregation, and 2D convolutions, are essential for our approach's effectiveness in enhancing scene understanding.

\label{sec:features}
\textbf{Volumetric local features:} We back project local feature map~$F_{I}$ along every ray to the world grid~($V$) to get 3D feature volume feature ($V_{F}$) with dimensions $K \times\ K  \times K \times 512$ where K is the resolution of feature grid and we use $K=64$. Note that there is a tradeoff between the size of the feature grid i.e. expressivity and computational cost. We found $K=64$ to give a reasonable performance while avoiding any OOM (out-of-memory issues) due to a larger grid size in our network training. \\
\textbf{Feature depth-encoding:} As detailed in Sec.~\ref{sec:triplanar}, we learn the depth of each feature in the feature grid using an additional 2-layer MLP with hidden dimension 512 to output depth-encoded features~$V_{Z}$ of dimensions~$K \times\ K  \times K \times 512$. Our feature aggregation module comprises three 2-layer MLPs with hidden dimension 512 and outputs learned weights $w_{i}$ over individual volumetric feature dimensions to outputs weights~$w_{xy}, w_{xz}$ and $w_{yz}$ each with dimensions $K \times  K \times K \times 1$. After performing $softmax$ and  summing over the $z, y$ and $x$ dimensions respectively, we obtain 2D feature maps for each of the three planes with dimension~$K \times K \times 512$. \\
\textbf{2D Convolutions}: We further use a series of 2D convolutions with upsampling layers to transform the planar features to dimension~$ H/2 \times W/2 \times 128$. The convolutional layers comprise three convolutional layers with input channels 512, 256, and 128 and output channels 256, 128, and 128 respectively, with a kernel size of 3, a stride of 2, and a padding of 1, followed by an upsampling layer with a scale factor of 2 and another convolutional layer with an input channel and output channel of $128$. Finally, an upsampling layer with an output dimension of $H/2 \times W/2$ is employed before outputting the features with a final convolution layer with input and output channels $128$. All convolutional layers are followed by the BatchNorm and ReLU layers. The output of each convolutional block becomes our tri-planar features~$S$, each with dimension~$ 128 \times 120 \times 160$. We sample into each plane in~$S$ by projecting $x$ into each plane~i.e. by getting the absolute $xy, xz$ and $yz$ coordinates of $x$ before concatenating and summing over the channel dimension to retrieve feature $f_{tp}$ with dimension $N \times128$ where N denotes the number of sampled points and 128 is the feature dimension. The residual local feature~$f_{r}$ after sampling into $F_{I}$ has dimensions $N \times 512$.

\subsection{Rendering MLPs:}
\label{sec:rendering_mlps}
The rendering MLPs for both foreground and background rendering comprises 7 fully-connected layers with hidden dimension of 128 and ReLU activation. We apply positional encoding~\cite{mildenhall2020nerf} to the input positions $x$ and viewing direction~$d$. We concatenate positions~$x$ with triplanar features~$f_{tp}$ and residual features~$f_{r}$ as an input to the first layer of the MLP. We also supply the conditioning feature as a skip connection to the third layer in the MLP and mean pool the features along the viewing dimension in the forth MLP layer, if there is more than one image in the input. We found this pooling strategy to work better than pooling before the rendering stage~i.e. earlier on in the tri-planar construction stage~(\ref{sec:features}). In total, we use the first 4 layers to output features of dimension~$N \times 128$, before utilizing a final density MLP to output 1 channel value for every sampled point $N$. We further use two additional dedicated MLP layers with a hidden dimension of 128 to output a 3-channel color value for every sampled point~$N$, conditioned on the positionally-encoded viewing direction and the output of the fourth MLP layer.

\section{Implementation Details:}
\label{sec:implementation_details}

\textbf{Sampling rays}: We scale all samples in the dataset so that cameras lie inside a unit hemisphere and use near and far values of 0.02 and 3.0 respectively. We use 64 coarse and 64 fine samples to sample each ray. 

\textbf{Training procedure}: To optimize~\titleShort{}, we first sample 3 source images from one of the 75 scenes in the training dataset. For our initial training phase, we sample 20 random destination views different from the source images used for encoding the~\titleShort{}'s network. We sample 1000 rays from all 20 destination views. We use these randomly sampled rays to decode the color and density for each of the 1000 rays. This training strategy helps the network simultaneously decode from a variety of camera distributions and helps with network convergence. We do this by sampling two different sets of points~i.e. one for each near and far background MLP, as employed in~\cite{zhang2020nerf++}. These points samples differ based on the intersection between the origin of rays and the unit sphere. 

\textbf{Loss function and optimizer:} For the first training phase, we employ a mean squared error loss on predicted color and target pixels at the sampled point locations in the ground-truth images, as discussed in Sec.~\ref{sec:decoding}. We also add a regularization penalty~(Eq.~\ref{eq:regularization}) to encourage the weights to be sparse for both near and far background MLPs, as proposed in~\cite{barron2021mip}. For our second training phase, we select a single destination view and sample~$40 \times 40$ patches of target RGB for training the network using an additional perceptual similarity loss, as described in Sec.~\ref{sec:decoding} with a~$\lambda$ value set to 0.3.${L}_{LPIPS}$ loss encourages perceptual similarity between patches of rendered color, $c^{p}$ and ground color~$\tilde{c}_t$, where we only enforce it after 30 training epochs to improve background modeling. We optimize the network for 100 epochs in total and employ early stopping based on the validation dataset which is a subset of the training dataset with different viewpoints than the training camera distribution. We use an Adam optimizer with an initial learning rate of~$5.0e^{-4}$ and a learning-rate ramp-up strategy to increase the learning rate from~$5.0e^{-5}$ to the value~$5.0e^{-4}$ and then decrease it exponentially to a final learning rate~$5.0e^{-6}$. 

\begin{strip}
\begin{center}
\centering
\includegraphics[width=1.0\textwidth]{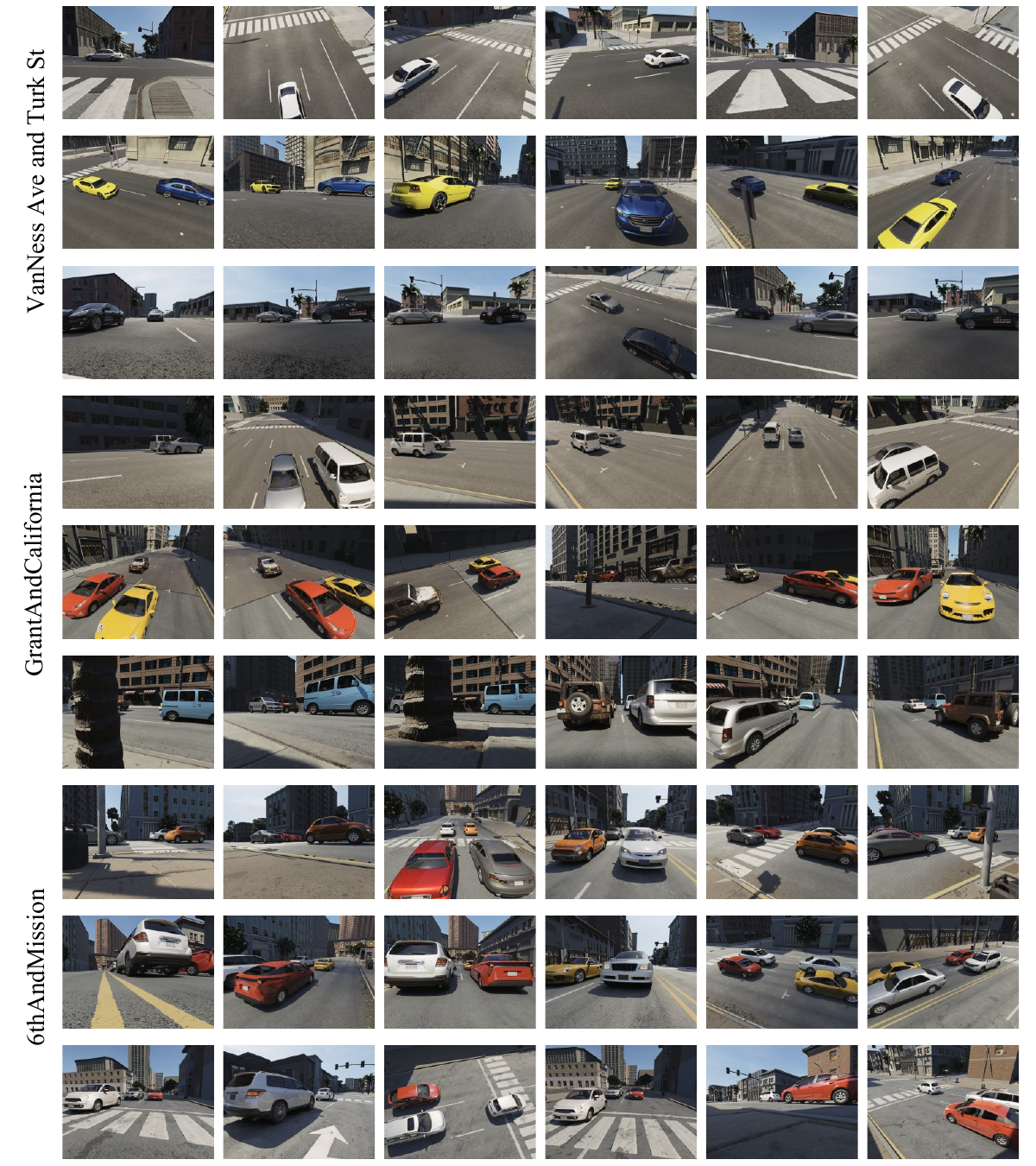}
\captionof{figure}{
\textbf{NeRDS360 training samples:} {We show diverse training samples from our proposed multi-view dataset, showing 3 examples for each map~(shown on the~$y$-axis) and displaying 5 randomly sampled images~(shown on the~$x$-axis) for each scene, from 100 rendered images with cameras placed in a hemisphere at a fixed radius from the center of the scene.
}}
\vspace{1.0cm}
\label{fig:train_samples}
\end{center}
\end{strip}

\begin{strip}
\begin{center}
\centering
\includegraphics[width=1.0\textwidth]{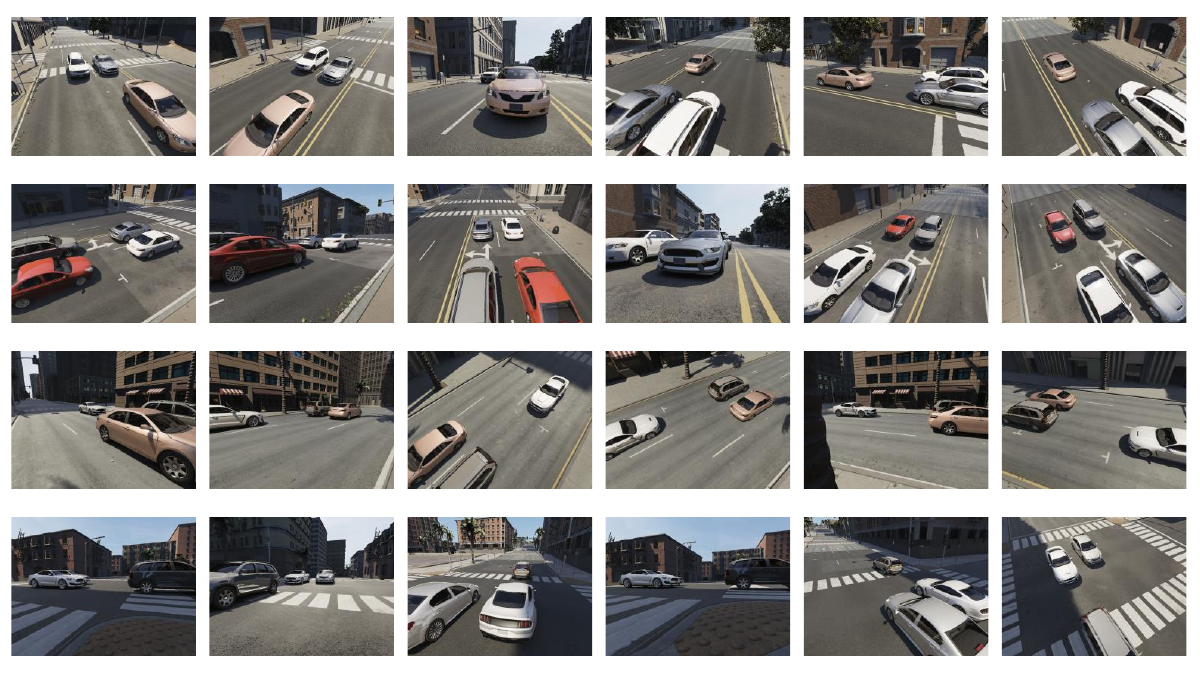}
\captionof{figure}{
\textbf{NeRDS360 test samples:} {We show unseen test samples with completely different backgrounds and objects not seen during training. Test samples include completely different camera viewpoints that are not observed during training which are still sampled in a hemisphere around the foreground objects of interest. Here, we show 4 different scenes from our evaluation dataset, different from the training dataset~(shown on the~$y$-axis) and show 5 randomly sampled images~(shown on the~$x$-axis) for each scene, from 100 rendered images with cameras placed in a hemisphere at a fixed radius from the center of the scene.
}}
\label{fig:test_samples}
\end{center}
\end{strip}

\textbf{Compute:} We train the model end-to-end on 8 A-100 Nvidia GPUs for approximately 1 day for network convergence. 

\textbf{Parameters:} Since~\titleShort{} has the ability to overfit to a large number of scenes, unlike NeRF~\cite{mildenhall2020nerf}, we use a larger model size of 17M parameters. Both ours and NeRF~\cite{mildenhall2020nerf}'s rendering MLP size is the same~(i.e. 1.2M parameters), although our larger model size is attributed to employing ResNet feature block for local features~($\sim$10M parameters) and additional convolutional blocks for tri-planar feature. 

\textbf{Optional fine-tuning:} Although our network gives reasonable zero-shot performance, we also employ an additional finetuning stage using the same few views~(e.g. 1, 3 and 5 source views) to further improve the performance of our network. Note that we employ the same finetuning strategy for the comparing baselines~(cf. Sec.~\ref{sec:experiments} in the main paper) and show that the additional finetuning stage improves the performance of both our proposed method and competing baseline, while our approach,~\titleShort{}, still achieves superior overall performance. For our finetuning experiments, we freeze the rest of the network and only the optimize tri-planar network i.e. freezing the encoder~$E$. We employ a lower learning rate of~$5^{10^{-6}}$ to finetune the network from 1,3 or 5 source views.

\begin{equation}
\label{eq:regularization}
\begin{array}{r}
\mathcal{L}_{\mathrm{reg}}(s, w)=\sum_{i=0}^{N-1} \sum_{j=0}^{N-1} w_i w_j|\frac{s_i+s_{i+1}}{2}-\frac{s_j+s_{j+1}}{2}| \\
+\frac{1}{3} \sum_{i=0}^{N-1} w_i^2 (s_{i+1}-s_i)
\end{array}
\end{equation}
\section{Experimental Setting Details:}
\label{sec:exp_details}
In this section, we detail our experimental setting to evaluate the effectiveness of our proposed method against the state-of-the-art baselines on the NeRDS360 dataset. We mainly evaluate for~\textbf{a.} Prior-based sampling and~\textbf{b.} Novel-scene rendering. Note that unlike~\cite{bautista2022gaudi} which performs both unconditional and conditional prior-based sampling, our task only considers image-conditional prior-based sampling for~\textbf{a}, since we don't optimize a latent code for each scene and our method doesn't rely on inference-time GAN-inversion like~\cite{bautista2022gaudi} to find a latent code for a new scene. Rather, our method works in a zero-shot manner reasonably well without any inference time finetuning or inversion, since it takes as input one or few images or a novel scene and is trained as such. We now describe more details about each experimental setting. ~\textbf{a. Prior-based sampling} tests for our network's ability to overfit the training distribution of a large number of scenes. In essence, we keep the evaluation scenes fixed to one of the scenes seen during training and use 1,3, and 5 source camera views as input while decoding from novel camera viewpoints not seen during training. While vanilla NeRF~\cite{mildenhall2020nerf} can do this with many different networks, each optimized from scratch from 100s of views for a new scene, our proposed approach, thanks to its generalizability can overfit to a large number of scenes with just a single network without optimizing a different latent code or vector per-scene, hence demonstrating our network's ability to memorize the training distribution for a large number of scenes seen during training.~\textbf{b. Novel-scene rendering} considers evaluating our approach on a completely new set of scenes and objects never seen during training. We test for our model's ability to generalize well in this scenario which is a core aspect of our approach. This is a more challenging evaluation setup than prior-based sampling since the network has not seen any scenes or objects, neither it has seen these viewpoints during training. Rather, it only relies on the priors learned during training and the few views available during testing~(1, 3, or 5 views in our evaluation setup) to infer the complete 360$^{\circ}$ surroundings of novel scenes.
\section{NeRDS360 Dataset:}
\label{sec:dataset_details}

In this section, we discuss our proposed NeRDS360 dataset. We show qualitative examples of our proposed dataset in Fig.~\ref{fig:train_samples} and Fig.~\ref{fig:test_samples}. Fig.~\ref{fig:train_samples} displays training samples of 3 different scenes from each of the 3 different maps in our dataset. Our dataset is very diverse both in terms of the scenes represented and the foreground car shapes and textures. NeRDS360's scenes also depict high variety in terms of occlusion of foreground objects~(i.e. not all foreground cars are observed from all views and there are various occluders such as trees and lightning poles present in the scene), varied number of objects represented~(i.e. we sample from 1 to 4 foreground cars for each scene with various textures, lightning, and shadows) as well as varied lighting and shadows in a scene~(i.e. lightning and shadows in each scene is not constant). Hence, our dataset and the corresponding task are extremely challenging. We also show different testing samples in Fig.~\ref{fig:test_samples}. As shown in the figure, we render completely novel viewpoints not seen during training as well as different textures and shapes of cars that are also not rendered during training. We evaluate for all 100 evaluation cameras sampled inside the hemisphere, as shown in Fig.~\ref{fig:testcameras} in the main paper while giving as input 1, 3, or 5 source views to the network.   
\section{Additional Qualitative Results:}
\label{sec:add_qualitative_results}
\begin{figure}[t]
\includegraphics[width=1.0\linewidth]{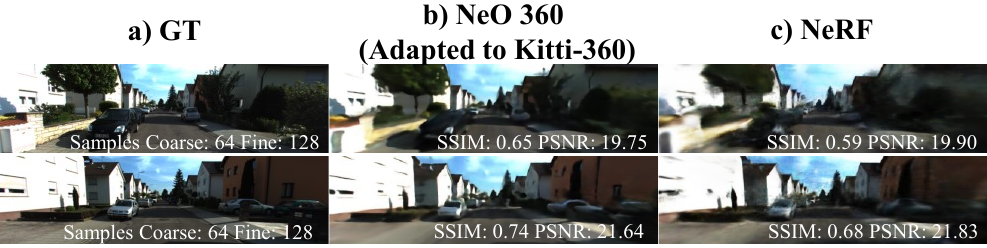}
\captionof{figure}{\textbf{Real-world results:} on KITTI-360~\cite{Liao2022PAMI}, Panoptic NeRF~\cite{fu2022panoptic} test split}
\vspace{-10pt}
\label{fig:KITTI-360}
\end{figure}

\looseness=-1
\textbf{Results on Kitti-360:} To demonstrate its applicability to real-world data, it's important to capture a comparable dataset to NeRDS360 in a real-world context. While our approach is tailored for a 360$^\circ$ environment, we have successfully adapted NeO 360 for the KITTI-360~\cite{Liao2022PAMI} dataset. This adaptation involves removing the distinction between near and far and employing a single MLP for rendering. Our method employs a source view window of the last 3 frames to render the subsequent frame. Examining overfitting outcomes~(Fig.~\ref{fig:KITTI-360}), we observe that our representation achieves significantly improved SSIM and comparable PSNR in contrast to NeRF, when dense views are available for real-world unbounded scenes.

%------------------------------------------------------------------------

%%%%%%%%% REFERENCES

\end{document}